\documentclass[11pt,letterpaper]{article}
\usepackage{naaclhlt2013}
\usepackage{times}
\usepackage{latexsym}
\usepackage{url}
\usepackage[utf8x]{inputenc}
\usepackage[pdftex]{graphicx}
\usepackage{amsmath}
\usepackage{color}
\usepackage{tabularx}
\usepackage{booktabs}
\usepackage{array}
\usepackage{amsmath}
\usepackage{xcolor,colortbl}
\definecolor{light-gray}{gray}{0.9}

\title{Unsupervised Induction of Contingent Event Pairs from Film Scenes}

\author{Zhichao Hu, Elahe Rahimtoroghi, Larissa Munishkina, Reid Swanson and Marilyn A. Walker \\
Natural Language and Dialogue Systems Lab \\
Department of Computer Science, 
  University of California, Santa Cruz \\
  Santa Cruz, CA, 95064 \\
  {\tt \{zhu, elahe, mlarissa, reid, maw\}@soe.ucsc.edu}}

\date{}

\begin{document}
\maketitle
\begin{abstract}
Human engagement in narrative is partially driven by reasoning about
discourse relations between narrative events, and the expectations
about what is likely to happen next that results from such reasoning.
Researchers in NLP have tackled modeling such expectations from a
range of perspectives, including treating it as the inference of the
{\sc contingent} discourse relation, or as a type of common-sense
causal reasoning. Our approach is to model likelihood between events by
drawing on several of these lines of previous work.  We implement and
evaluate different unsupervised methods for learning event pairs that
are likely to be {\sc contingent} on one another. We refine event pairs that we learn from a corpus of
film scene descriptions utilizing web search counts, and evaluate
our results by collecting human judgments of contingency.
Our results indicate that the use of web search
counts increases the average accuracy of our best
method to 85.64\% 
over a baseline of 50\%,
as compared to an average accuracy of 75.15\% 
without web search.
\end{abstract}

\section{Introduction}
\label{intro-sec}

Human engagement in narrative is partially driven by reasoning about
discourse relations between narrative events, and the expectations about
    what is likely to happen next that results from such reasoning
\cite{Gerrig83,Graesseretal94,Lehnert81,Goyaletal10}.  Thus discourse
relations are one of the primary means to structure
narrative in genres as diverse as weblogs, search queries, stories, film scripts and
news articles
\cite{ChambersJurafsky09,Manshadietal08,GordonSwanson09,Gordonetal11,BeamerGirju09,RiazGirju10,DoChaRo11}.

\begin{figure}[ht]
  \centering
	\small
  \begin{tabular}{|p{3.0in}|}
\hline
 DOUGLAS QUAIL and his wife KRISTEN, are asleep in bed. \\

Gradually the room lights brighten.  the clock chimes and
        begins speaking in a soft, feminine voice.   \\
They don't budge.  Shortly, the clock chimes again. \\
Quail's wife stirs.  Maddeningly, the clock chimes a third time. \\
CLOCK (continuing)Tick, tock --. \\
Quail reaches out and shuts the clock off.  Then he sits up in bed. \\
He swings his legs out from under the covers and sits on the edge of the bed.  He puts on his glasses and sits, lost in thought. \\
He is a good-looking but conventional man in his early thirties.  He seems rather in awe of his wife, who is attractive and rather off-hand towards him. \\
Kirsten pulls on her robe, lights a cigarette, sits fishing for her slippers. \\
\hline
    \end{tabular}
  \caption{Opening Scene from Total Recall}
  \label{quaid-fig-new}
\end{figure}

Recent work in NLP has tackled the inference of relations between
events from a broad range of
perspectives: (1) as inference of a discourse relations (e.g. the Penn Discourse Treebank (PDTB)
{\sc contingent} relation and its specializations); (2) as a type of
common sense reasoning; (3) as part of text understanding to support
question-answering; and (4) as way of learning script-like or
plot-like knowledge structures. All these lines of work aim to model
narrative understanding, i.e. to enable systems to infer which events
are likely to {\bf have happened} even though they have not been
mentioned in the text \cite{Schank77}, and which events are likely {\bf to
happen} in the future. Such knowledge has practical applications 
in commonsense reasoning, information retrieval, question answering, 
narrative understanding and inferring discourse relations.


We model this likelihood between events by drawing on the PTDB's
general definition of the {\sc contingent} relation, which
encapsulates relations elsewhere called {\sc cause}, {\sc condition}
and {\sc enablement}
\cite{Prasadetal08,Linetal10,Pitleretal09,Louisetal10}.  Our aim in
this paper is to implement and evaluate a range of different
unsupervised methods for learning event pairs that are likely to be
{\sc contingent} on one another.

We first utilize a corpus of scene descriptions from films because
they are guaranteed to have an explicit narrative structure.  

Screenplay scene descriptions are one type of
narrative that tend to be told in temporal order
\cite{BeamerGirju09,GordonSwanson09}, which makes them a good resource
for learning about contingencies between events.  In addition, scenes
in film represent many typical sequences from real life while providing
a rich source of event clusters related to battles, love and mystery.
We carry out separate experiments for the action movie genre and the romance
movie genre.
For example, in the scene from {\it
  Total Recall}, from the action movie genre (See
Fig.~\ref{quaid-fig-new}), we might learn that the event of {\tt
  sits up} is {\sc contingent} on the event of {\tt clock chimes}. The
subset of the corpus we use comprises a total of 123,869 total unique
event pairs.

We produce initial scalar estimates of potential {\sc contingency}
between events using four previously defined measures of
distributional co-occurrence. We then refine these estimates through
web searches that explicitly model the patterns of narrative event
sequences that were previously observed to be likely within a particular
genre. There are several advantages of this method: (1) events in the
same genre tend to be more similar than events across genres, so less
data is needed to estimate co-occurrence; (2) film scenes are
typically narrated via simple tenses in the correct temporal order,
which allows the ordering of events to contribute to the estimation of
the {\sc contingency} relation; (3) The web counts focus on validating
event pairs already deemed to be likely to be {\sc contingent} in the
smaller, more controlled, film scene corpus. To test our method, we
conduct perceptual experiments with human subjects on Mechanical Turk
by asking them to select which of two pairs of events are the most
likely. For example, given the scene from {\it Total Recall} in
Fig.~\ref{quaid-fig-new}, Mechanical Turkers are asked to select
whether the sequential event pair {\tt clock chimes, sits up} is more
likely than {\tt clock chimes} followed by a randomly selected event
from the {\bf action} film genre.  Our experimental data and
annotations are available at
\url{http://nlds.soe.ucsc.edu/data/EventPairs}.

Sec.~\ref{exp-sec} describes our experimental method in
detail. Sec.~\ref{eval-results-sec} describes how we set up our
evaluation experiments and the results.  We show that none of the methods
from previous work perform better on our data than 75.15\% average
accuracy as measured by human perceptions of {\sc contingency}.  But
after web search refinement, we achieve an average accuracy
of 85.64\%.
We delay a more detailed comparison to previous work to
Sec.~\ref{rel-work-sec} where we summarize our results
and compare previous work to our own.


\section{Experimental Method}
\label{exp-sec}

Our method uses a combination of estimating the
likelihood of a {\sc contingent} relation between events in a corpus
of film scenes \cite{Walkeretal12d}, with estimates then revised
through web search. Our experiments are based on two subsets of 862
film screen plays collected from the IMSDb website using its ontology
of film genres \cite{Walkeretal12d}: a set of {\bf action} movies of
115 screenplays totalling 748 MB, and a set of {\bf romance} movies of
71 screenplays totalling 390 MB.  Fig.~\ref{quaid-fig-new} provided an
example scene from the action movie genre from the IMSDb corpus.

We assume that the relation we are aiming to learn is the PDTB {\sc
  contingent} relation, which is defined as a relation that exists
when one of the situations described in the text spans that are
identified as the two arguments of the relation, i.e. Arg1 and Arg2,
causally influences the other \cite{prasad-etal-lrec08}.  As Girju
notes, it is notoriously difficult to define causality without making
the definition circular, but we follow Beamer and Girju's work in
assuming that if events A, B are causally related then B should occur
less frequently when it is not preceded by A and that B $\rightarrow$A
should be much less frequent than A $\rightarrow$ B.  We assume that
both the {\sc cause} and {\sc condition} subtypes of the {\sc
  contingency} relation will result in pairs of events that are likely
to occur together and in a particular order.  In particular we assume
that the subtypes of the PDTB taxonomy of Contingency.Cause.Reason and
Contingency.Cause.Result are the most likely to occur together as
noted in previous work.  Other related work has made use of discourse
connectives or discourse taggers (implicit discourse relations) to
provide additional evidence of {\sc contingency}
\cite{DoChaRo11,Gordonetal11,Chiarcos12,Pitleretal09,Linetal10}, but
we do not because the results have been mixed. In particular these
discourse taggers are trained on The Wall Street Journal (WSJ) and are
unlikely to work well on our data.

We define an event as a verb lemma with its subject and object. Two
events are considered equal if they have the same verb.  We do not
believe word ambiguities to be a primary concern, and previous work
also defines events to be the same if they have the same surface verb,
in some cases with a restriction that the dependency relations should
also be the same
\cite{ChambersJurafsky08,ChambersJurafsky09,DoChaRo11,RiazGirju10,Manshadietal08}.
Word sense ambiguities are also reduced in specific genres (Action and
Romance) of film scenes.

Our method for estimating the likelihood of a {\sc contingent} relations
between events consists of four steps:
\begin{enumerate}
\item {\sc text processing}: We use Stanford
CoreNLP to annotate the corpus document by document
and stored the annotated text in XML format (Sec.~\ref{tp-sec});
\item {\sc compute event representations}: Form intermediate artifacts
such as events,  protagonists and
event pairs from the annotated documents. Each
event has its arguments (subject and object). We calculate the
frequency of the event across the relevant genre (Sec.~\ref{eventrep-sec});
\item {\sc calculate contingency measures}: We define 4 different
measures of contingency and calculate each one separately using
the results from Steps 1 and 2 above. We call each result a {\sc predicted causal event pair} ({\bf PCEP}). All measures return scalar values that
we use to rank the PCEPs  (Sec.~\ref{measure-sec});
\item {\sc web search refinement}: We select the top 100 event pairs
  calculated by each contingency measure, and construct a {\sc random
    event pair} ({\bf REP}) for each PCEP that preserves the first
  element of the PCEP, and replaces the second element with another event
  selected randomly from within the same genre.  We then define web
  search patterns for both PCEP and REPs and compare the counts
  (Sec.~\ref{web-search-sec}).
\end{enumerate}

\subsection{Text Processing}
\label{tp-sec}

We first separate our screen plays into two sets of documents, one for
the action genre and one for the romance genre. Because we are
interested in the event descriptions that are part of the scene
descriptions, we excise the dialog from each screen play.  Then using
the Stanford CoreNLP pipeline, we annotate the film scene
files. Annotations include tokenization, lemmatization, named entity
recognition, parsing and coreference resolution.

We extract the events by keeping all tokens whose POS tags begin with
{\tt VB}. We then use the dependency parse to find the subject and
object of each verb (if any), considering only {\tt nsubj, agent,
  dobj, iobj, nsubjpass}.  We keep the original tokens of the subject
and the object for further processing.

\subsection{Compute Event Representations}
\label{eventrep-sec}

Given the results of the previous step we start by generalizing the
subject and object stored with each event by substituting tokens with
named entities if there are any named entities tagged.  Otherwise we
generalize the subjects and the objects using their lemmas. For
example, {\tt person UNLOCK door}, as illustrated in Table~\ref{ws-res}.

We then integrate all the subjects and objects across all film scene
files, keeping a record of the frequency of each subject and
object. For example, {\tt [person (115), organization (14), door (3)]
  UNLOCK [door (127), person (5), bars (2)]}. The most frequent
subject and object are selected as representative arguments for the
event.  We then count the frequency of each event across all the film
scene files.

Within each film scene file, we count adjacent events as potential
{\sc contingent} event pairs. Two event pairs are defined as equal if
they have the same verbs in the same order. We also count the frequency of
each event pair.

\subsection{Calculate Contingency Measures}
\label{measure-sec}

We calculate four different measures of {\sc contingency} based on
previous work using the results of Steps 1 and 2 (Sec.~\ref{tp-sec}
and Sec. ~\ref{eventrep-sec}).  These measures are pointwise mutual
information, causal potential, bigram probability and
protagonist-based causal potential as described in detail below. We
calculate each measure separately by genre for the action and romance
genres of the film corpus. \\

\noindent{\bf Pointwise Mutual Information.} The majority of 
related work uses pointwise mutual information ({\bf
  PMI}) in some form or another
\cite{ChambersJurafsky08,ChambersJurafsky09,RiazGirju10,DoChaRo11}.
Given a set of events (a verb and its collected set of subjects and
objects), we calculate the PMI using the standard definition:

\begin{equation}
\mathit{pmi}(e_1, e_2) = \log\frac{P(e_1, e_2)}{P(e_1)P(e_2)}
\end{equation}

in which $e_1$ and $e_2$ are two events. $P(e_1)$ is the probability that event $e_1$ occur in the corpus: 

\begin{equation}
P(e_1) = \frac{count(e_1)}{\sum_x count(e_x)}
\end{equation}

where $count(e_1)$ is the count of how many times event $e_1$ occurs in the corpus, and $\sum_x count(e_x)$ is the count of all the events in the corpus. The numerator is the probability that the two events occur together in the corpus:
 \begin{equation}
 P(e_1, e_2) = \frac{count(e_1, e_2)}{\sum_x\sum_y count(e_x, e_y)}
 \end{equation}

 in which $count(e_1, e_2)$ is the number of times the two events
 $e_1$ and $e_2$ occur together in the corpus regardless of their
 order. Only adjacent events in each document are paired up. PMI is a
 symmetric measurement for the relationship between two events. The
 order of the events does not matter. \\

\noindent{\bf Causal Potential.} Beamer and Girju proposed a measure
called Causal Potential ({\bf CP}) based on previous work in
philosophy and logic, along with an annotation test for causality.  An
annotator deciding whether event A causes event B asks herself the
following questions, where answering yes to both means the two events
are causally related:
\begin{itemize}
\item  Does event A occur
before (or simultaneously) with event B? 
\item  Keeping constant as many
other states of affairs of the world in the given text context as
possible, does modifying event A entail predictably modifying event B?
\end{itemize}

As Beamer \& Girju note, this annotation test is objective, and it is
simple to execute mentally. It only assumes that the
average person knows a lot about how things work in the world and can
reliably answer these questions. 
{\bf CP} is then defined below, where the arrow notation means ordered bigrams, i.e. event $e_1$ occurs before event $e_2$:

\begin{equation}
\phi(e_1, e_2) = \mathit{pmi}(e_1, e_2) + \log\frac{P(e_1\rightarrow e_2)}{P(e_2\rightarrow e_1)}\\
\label{eq:cp}
\end{equation}
\begin{equation*}
\mathrm{where}~\mathit{pmi}(e_1, e_2) = \log\frac{P(e_1, e_2)}{P(e_1)P(e_2)}
\end{equation*}

The causal potential consists of two terms: the first is pair-wise
mutual information (PMI) and the second is relative ordering of
bigrams. PMI measures how often events occur as a pair; whereas
relative ordering counts how often event order occurs in the
bigram. If there is no ordering of events, the relative ordering is
zero. We smooth unseen event pairs by setting their
frequency equal to 1 to avoid zero probabilities. For {\bf CP} as with PMI,
we restrict these calculations to adjacent events. Column CP of
Table~\ref{ws-res} below provides sample values for the CP measure.\\

\begin{table*}[ht]
\centering
\begin{scriptsize}
\begin{tabular}{|r|p{2.9cm}|p{.35cm}|p{2.1cm}|r|p{2.8cm}|p{2.1cm}|r|}
\hline
Row \#&Causal Potential Pair & CP & PCEP Search pattern & NumHits & Random Pair & REP Search pattern & NumHits  \\ \hline \hline

1&person KNOW person - person MEAN what & 2.18 &  he knows * means  &  415M  & person KNOW person - person PEDDLE papers &  he knows * peddles  & 2  \\ \hline
2&person COME - person REST head & 2.12 &  he comes * rests  &  158M  & person COME - person GLANCE window &  he comes * glances  & 41  \\ \hline
3&person SLAM person - person SHUT door & 2.11 &  he slams * shuts  & 11 & person SLAM person - person CHUCKLE &  he slams * chuckles  & 0  \\ \hline
4&person UNLOCK door - person ENTER room & 2.11 &  he unlocks * enters  & 80 & person UNLOCK door - person ACT shot &  he unlocks * acts  & 0  \\ \hline
5&person SLOW person - person STOP person & 2.10 &  he slows * stops  &  697K  & person SLOW person - eyes RIVET eyes &  he slows * rivets  & 0 \\ \hline
6&person LOOK window - person WONDER thing & 2.06 &  he looks * wonders  &  342M  & person LOOK window - person EDGE hardness &  he looks * edges  & 98  \\ \hline
7&person TAKE person - person LOOK window & 2.01 &  he takes * looks  &  163M  & person TAKE person - person CATCH person &  he takes * catches  &  311M \\ \hline
8&person MANAGE smile - person GET person & 2.01 &  he manages * gets  &  80M  & person MANAGE smile - person APPROACH person &  he manages * approaches  & 16 \\ \hline
9&person DIVE escape - person SWIM way & 2.00 &  he dives * swims  &  1.5M  & person DIVE escape - gun JAM person &  he dives * jams  & 6  \\ \hline
10&person STAGGER person - person DROP person & 2.00 &  he staggers * drops  & 33 & person STAGGER person - plain WHEEL person &  he staggers * wheels  & 1 \\ \hline
11&person SHOOT person - person FALL feet & 1.99 &  he shoots * falls  &  55.7M  & person SHOOT person - person PREVENT person &  he shoots * prevents  & 6 \\ \hline
12&person SQUEEZE person - person SHUT door & 1.87 &  he squeezes * shuts  & 5 & person SQUEEZE person - person MARK person &  he squeezes * marks  & 1  \\ \hline
13&person SEE person - person GO & 1.87 &  he sees * goes  &  184M  & person SEE person - image QUIVER hips &  he sees * quivers  & 2 
\\ \hline
\end{tabular}
\caption{\label{ws-res} Sample web search patterns and values used in web search refinement algorithm from action genre}
\end{scriptsize}
\end{table*}

\noindent{\bf Probabilistic Language Models.}  
Our third method models event sequences using statistical
language models ~\cite{Manshadietal08}. A language model estimates
the probability of a sequence of words using a sample corpus.
To identify contingent event sequences, we apply a bigram model which
estimates the probability of observing the sequence of two words $w_1$
and $w_2$ as follows:

\begin{equation}
P(w_1, w_2) \cong P(w_2 | w_1) = \frac{count(w_1, w_2)}{count(w_1)}
\label{eq:bigram}
\end{equation}

Here, the words are events. Each verb is a single event and each film
scene is treated as a sequence of verbs. For example, consider the
following sentence from \textit{Total Recall}:

\begin{quotation}
\em Quail and Kirsten sit at a small table, eating breakfast.
\end{quotation}

This sentence is represented as the sequence of its two verbs: {\tt
  sit, eat}.  We estimate the probability of verb bigrams using
Equation~\ref{eq:bigram} and hypothesize that the verb sequences with
higher probability are more likely to be contingent. We apply a
threshold of 20 for $count(w_1, w_2)$ to avoid infrequent and uncommon
bigrams. \\

\noindent{\bf Protagonist-based Models.}  We also used a method of
generating event pairs based not only on the consecutive events in
text but on their protagonist. This is based on the assumption that
the agent, or protagonist, will tend to perform actions that further
her own goals, and are thus causally related.  We called this method
protagonist-based because all events were partitioned into multiple
sets where each set of events has one protagonist. This method is roughly
based on previous work using chains of discourse entities to induce narrative
schemas \cite{ChambersJurafsky09}.

Events that share one protagonist were extracted from text according
to co-referring mentions provided by the Stanford CoreNLP
toolkit.\footnote{http://nlp.stanford.edu/software/corenlp.shtml}  A
manual examination of coreference results on a sample of movie scripts
suggests that the accuracy is only around 60\%: most of the time the
same entity (in its nominal and pronominal forms) was not recognized
and was assigned as a new entity.

We preserve the order of events based on their textual order 
assuming as above that film scripts tend to preserve temporal order.
An ordered event pair is generated if both events
share a protagonist. We further filter event pairs by eliminating
those whose frequency is less than 5 to filter insignificant
and rare event pairs. This also tends to catch errors generated by the
Stanford parser.

{\bf CP} was then calculated accordingly to Equation~\ref{eq:cp}.  To
calculate the PMI part of {\bf CP}, we combine the frequencies of
event pairs in both orders.

\subsection{Web Search Refinement}
\label{web-search-sec}

We then define web search patterns based on the PCEPs 
that we have learned from the film corpus.
Recall that {\bf REP} stands for random event pair,
and that {\bf PCEP} stands for predicted contingent event pair.
Our hypothesis is that using the film corpus within a particular genre 
to do the initial estimates of contingency takes advantage of genre
properties such as similar events and narration of scenes in chronological
order. However the film corpus is necessarily small, and we can
augment the evidence for a particular contingent relation by defining
specific narrative sequence patterns and collecting web counts. 
PCEPs should be frequent in web search and REPs should be infrequent. 

Our web
refinement procedure is:
\begin{itemize}
\item  For each event pair, create a Google search item 
as illustrated by Table~\ref{ws-res}, and described in more detail below.
\item Search for the exact match in Google general web search using incognito 
browsing and record the estimated count of results returned; 
\item Remove all the PCEP/REP pairs
with CP Google search count less than 100: highly contingent events should
be frequent in a general web search;
\item Remove all PCEP/REP pairs with REP Google search count greater
than 100: events that are not contingent on one another should not be frequent in a general web search.
\end{itemize}

The motivation for this step is to provide additional evidence {\bf
  for} or {\bf against} the contingency of a pair of events.
Table~\ref{ws-res} shows a selection of the top 100 CPEPs learned
using the Causal potential ({\bf CP}) Metric, the web search patterns
that are automatically derived from the CPEPs (Column 4), the REPs
that were constructed for each CPEP (Column 6), the web search
patterns that were automatically derived from the REPs (Column 7).
Column 5 shows the results of web search hits for the CPEP patterns
and Column 8 shows the results of web search hits for the REP
patterns.  These hit counts were then used in refining our estimates
of {\sc contingency} for the learned patterns as described above.

Note that the web search patterns do not aim to find every possible
match of the targeted {\sc contingent} relation that could possibly
occur. Instead, they are generalizations of the instances of PCEPs
that we found in the films corpus. They are targeted at finding hits
that are the most likely to occur in {\bf narrative sequences}, which
are most reliably signalled by use of the historical present tense,
e.g. {\it He knows} in Row 1 and {\it He comes} in Row 2 of
Table~\ref{ws-res}.
\cite{SwansonGordon12,BeamerGirju09,LabovWaletzky97}. These search
patterns are not intended to match the original instances in the film
corpus and in general they are unlikely to match those instances.  In
addition, we use the ``*'' operator in Google Search to limit search
to pairs of events reported in the historical present tense, that are
``near'' one another, and in a particular sequence. We don't care
whether the events are in the same utterance or in sequential
utterances, thus for the second verb (event) we do not include a
subject pronoun {\it he}.

For example, consider the search patterns and results shown in Row 1
of Table~ \ref{ws-res}. The CPEP is {\tt person KNOW person, person
  MEAN what}. The REP is {\tt person KNOW person, person PEDDLE
  papers}. Our prediction is that the REP should be much less likely
in web search counts and the results validate that predication. A
paired t-test over the 100 top CPEP pairs for the CP measure comparing
the hit counts for the CPEP pairs vs. the REP pairs was highly
significant (p $<$ .00001). However, consider Row 7. Even though in
general the CPEP pairs are more likely (as measured by the web search
counts), there are cases where the REP is highly likely as shown by
the REP {\tt person take person, person CATCH person}) in Row
7. Alternatively there are cases where the web search counts provide
evidence against one of the PCEPs. Consider Rows 3, 4 10 and 12. In
all of these cases the web counts NumHits for the CPEP are in the
tens.

After the web search refinement, we retain the PCEP/REP 
pairs with initially high PCEP estimates, for which we found good
evidence for contingency and for randomness, e.g. Row 1 and 2
in Table~\ref{ws-res}.  We use 100 as a threshold because most of the
time the estimate result count from Google is either a very large
number (millions) or a very small number (tens), as illustrated by
the NumHits columns in Table~\ref{ws-res}.

We experimented with different types of patterns with a development
set of CPEPs before we settled on the search pattern template shown in
Table~\ref{ws-res}. We decided to use third person rather than first
person patterns, because first person patterns are only one type of
narrative \cite{SwansonGordon12}.  We also decided to utilize event
patterns without typical objects, such as {\it head} in {\tt person
  REST head} in Row 2 of Table~\ref{ws-res}.  We do not have any
evidence that this is the optimal search pattern template because
we did not systematically try other types of search patterns.

\begin{figure*}[htb]
\centering
\includegraphics[width=4.25in]{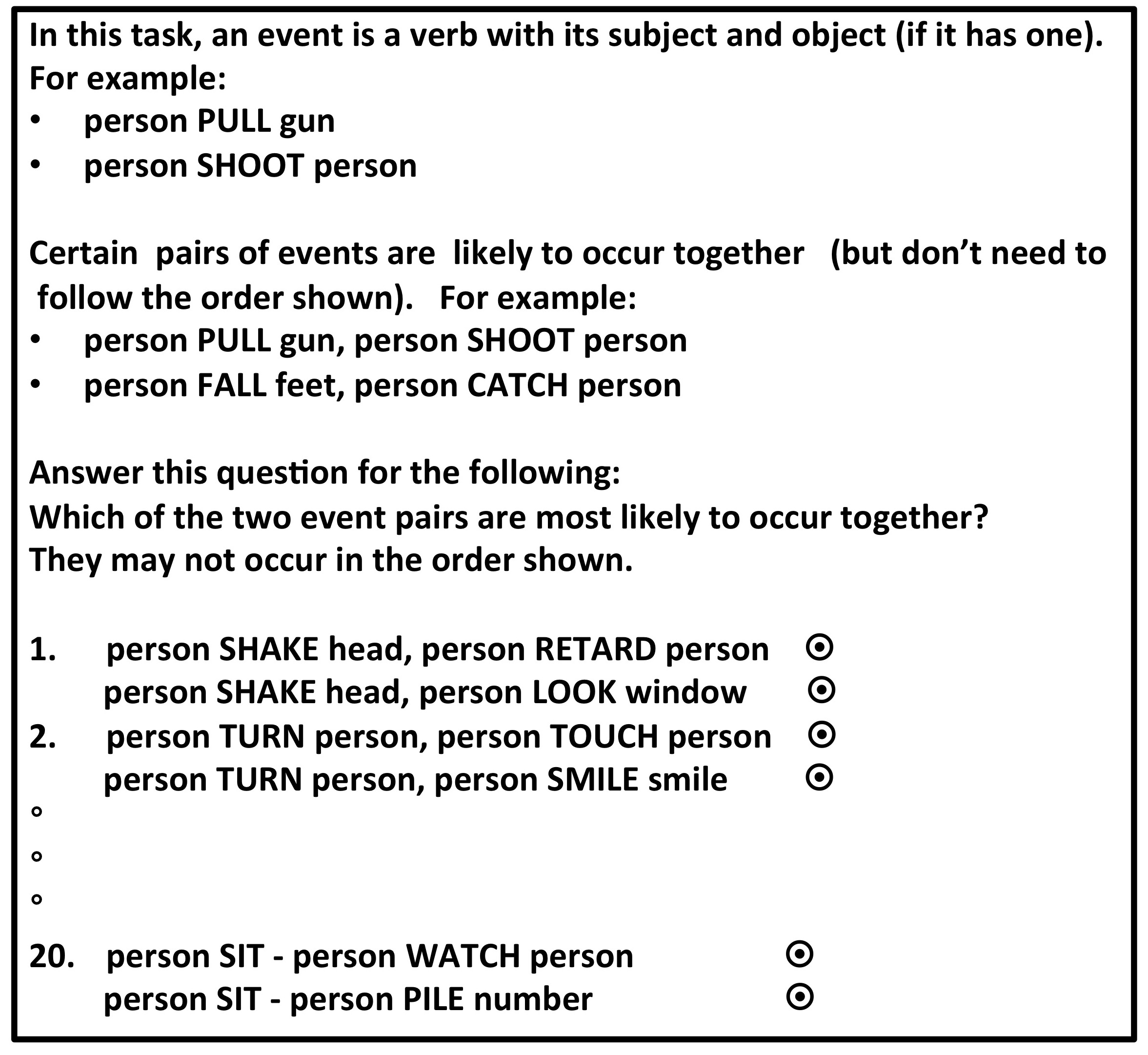}
\caption{\label{mturk-hit-args} Mechanical Turk HIT with event arguments
provided. This HIT also illustrates instructions where Turkers are told that the order of the events {\bf does not} matter.}
\end{figure*}

\section{Evaluation and Results}
\label{eval-results-sec}

While other work uses a range of methods for evaluating accuracy, to
our knowledge our work is the first to use human judgments from
Mechanical Turk to evaluate the accuracy of the learned contingent
event pairs. We first describe the evaluation setup in Sec.~\ref{mt-eval-sec}
and then report the results in Sec.~\ref{results-sec}

\subsection{Mechanical Turk Contingent Pair Evaluations}
\label{mt-eval-sec}

\begin{figure*}[htb]
\centering
\includegraphics[width=4.25in]{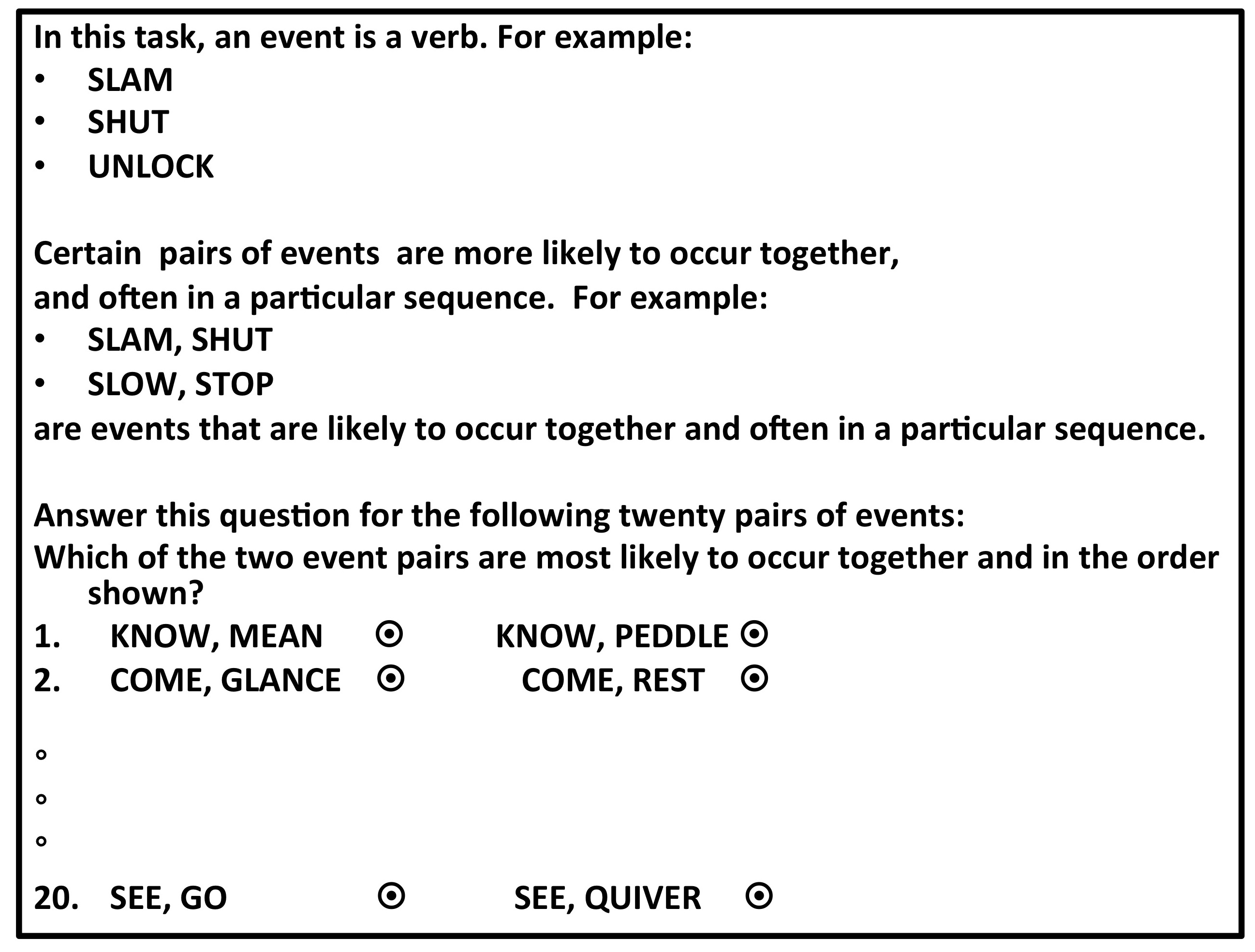}
\caption{\label{mturk-hit-noargs} Mechanical Turk HIT for
evaluation with no event arguments provided. This HIT also illustrates instructions where Turkers are told that the order of the events {\bf does} matter.}
\end{figure*}

We used three different types of HITs (Human Intelligence Tasks) on
Mechanical Turk for our evaluation.  Two of the HITS are in
Fig.~\ref{mturk-hit-args} and Fig.~\ref{mturk-hit-noargs}. The
differences in the different types of HITS involve: (1) whether the
arguments of events were given in the HIT, as in Fig.~\ref{mturk-hit-args}
and (2): whether the Turkers were told that the order of the events
mattered, as in Fig.~\ref{mturk-hit-noargs}. We initially thought that
providing the arguments to the events as shown in
Fig.~\ref{mturk-hit-args} would help Turkers to reason about which even
was more likely. We tested this hypothesis only in the action genre
for the Causal Potential Measure.  For CP, Bigram and Protag the order
of events always matters.  For the PMI task, the order of the events
doesn't matter because PMI is a symmetric
measure. Fig.~\ref{mturk-hit-args} illustrates the instructions that were
given with the HIT when the event order doesn't matter. In all the
other cases, the instructions that were given with the HIT are those
shown in Fig.~\ref{mturk-hit-noargs} where the Turkers are instructed to
pay attention to the order of the events given.

For all types of HITS, for all measures of {\sc contingency} we set up
the task as a choice over two alternatives, where for each predicted
contingent pair (PCEP), we generate a random event pair (REP), with
the first event the same and the second one randomly chosen from all
the events in the same film genre.  The REPs are constructed the same
way as we construct REPs for web search refinement, as illustrated by
Table~\ref{ws-res}.  This is illustrated in both Fig.~\ref{mturk-hit-args}
and Fig.~\ref{mturk-hit-noargs}. For all types of HITS, we ask 15 Turkers
from a pre-qualified group to select which pair (the PCEP or the REP)
are more likely to occur together. Thus, the framing of these
Mechanical Turk tasks only assumes that the average person knows how
the world works; we do not ask them to explicitly reason about causality
as other work does \cite{BeamerGirju09,Gordonetal11,DoChaRo11}.

For each measure of {\sc contingency}, we take 100 event pairs with
highest PCEP scores, and put them in 5 HITs with twenty items per HIT.
Previous work has shown that for many common NLP tasks, 7 Turkers'
average score can match expert annotations \cite{Snowetal08}, however
we use 15 Turkers because we had no gold-standard data and because we
were not sure how difficult the task is. It is clearly subjective.
Then to calculate the accuracy of each method, we computed the average
correlation coefficient between each pair of raters and eliminated the
5 lowest scoring workers. We then used the perceptions of the 10
remaining workers to calculate accuracy as \#correct answers
\\ \#total number of answers. 

In general, deciding when a MTurk worker is unreliable when the data
is subjective is a difficult problem. In the future we plan to test
other solutions to measuring annotator reliability as proposed in
related work
~\cite{callison-burch_fast_2009,Snowetal08,karger_iterative_2011,dawid_maximum_1979,welinder_multidimensional_2010,liu_variational_2012}.




\subsection{Results}
\label{results-sec}

We report our results in terms of overall accuracy.
Because the Mechanical Turk task is a choose-one question
rather than a binary classification, $\text{Precision} = \text{Recall}$
in our experimental results:
\begin{align*}
&\text{True Positive} = \text{Number of Correct Answers} \\
&\text{True Negative} = \text{Number of Correct Answers} \\
&\text{False Positive} = \text{Number of Incorrect Answers} \\
&\text{False Positive} = \text{Number of Incorrect Answers} \\
&\text{Precision} = \frac{\text{True Positive}}{\text{True Positive}+\text{False Positive}}\\
&\text{Recall} = \frac{\text{True Positive}}{\text{True Positive}+\text{False Negative}}
\end{align*}

\begin{table*}[!htb]
  \begin{center}
    \begin{tabular}{|p{.8cm}|p{6.0cm}|p{2.0cm}|p{2.25cm}|p{2.25cm}|}
    \hline
Row \#& Contingency Estimation Method & Action Acc\% &  Romance Acc\%  & Average Acc\%\\
\hline
\hline
1&CP with event arguments & 69.30\% & NA & 69.30\\
\hline
2 &CP with event arguments + Web search& 77.57\% & NA & 77.57 \\
\hline
3&CP no args  &  75.20  &  75.10  & 75.15\\
\hline
4&CP no args +Web Search & \bf{87.67 } &  83.61  & {\bf 85.64} \\
\hline \hline 
5&PMI  no args & 68.70 & 79.60 & 74.15\\
    \hline
6&PMI no args +Web Search & 72.11  &  \bf{88.52 } & 80.32\\
    \hline     \hline
7&Bigram no args & 67.10  &  66.50  & 66.80\\
    \hline
  8&  Bigram no args +Web Search & 72.40  & 70.14 & 71.27\\
    \hline     \hline
   9& Protag CP no args & 65.40  & 68.20 & 66.80\\
    \hline
   10& Protag CP no args +Web Search  & 76.59  & 64.10 & 70.35\\
    \hline
    \end{tabular}
  \end{center}
  \caption{\label{eval_all_methods} Evaluation results for the top 100 event pairs using all methods.}
\end{table*}

The accuracies of all the methods are shown in
Table~\ref{eval_all_methods}.  The results of using event arguments 
({\tt person KNOW person}) in
the Mechanical Turk evaluation task (i.e. 
Fig.~\ref{mturk-hit-args}) 
is given in
Rows 1 and 2 of Table~\ref{eval_all_methods}. The accuracies for Rows
1 and 2 are considerably lower than when the PCEPs are tested without
arguments.  Comparing Rows 1 and 2 with Rows 3 and 4 suggests that
even if the arguments provide extra information that help to ground
the type of event, in some cases these constraints on events may
mislead the Turkers or make the evaluation task more difficult.
There is an over 10\% increase in CP + Web search accuracy when we
compare Row 2 with Row 4.
Thus omitting the arguments of events in evaluations actually appears
to allow Turkers to make better judgments. 

In addition, Table~\ref{eval_all_methods} shows clearly that for every
single method, accuracy is improved by refining the initial estimates
of contingency using the narrative-based web search patterns.  Web
search increases the accuracy of almost all evaluation tasks, with
increases ranging from 3.45\% to 12.5\% when averaged over both film
genres (column 3).  The best performing method for the Action genre is
CP+Web Search at 87.67\%, while the best performing method for the
Romance genre is PMI+Web search at 88.52\%. However PMI+Web Search
does not beat CP+Web Search on average over both genres we tested,
even though the Mechanical Turk HIT for CP specifies that the order of
the events matters: a more stringent criterion. Also overall the
CP+WebSearch method achieves a very high 85.64\% accuracy.

It is also interesting to note the variation across the different
methods. For example, while it is well known that PMI typically
requires very large corpora to make good estimates, the PMI method
without web search refinement has an initially high accuracy of
79.60\% for the romance genre, while only achieving 68.70\% for
action.  Perhaps this difference arises because the romance genre is
more highly causal, or because situations are more structured in
romance, providing better estimates with a small corpus.  However even
in this case of romance with PMI, adding web search refinement
provides an almost 10\% increase in absolute accuracy to the highest
accuracy of any combination, i.e. 88.52\%.  There is also an
interesting case of Protag CP for the romance genre where web search
refinement actually decreases accuracy by 4.1\%.  In future work we
plan to examine more genres from the film corpus and also examine the
role of corpus size in more detail.




\section{Discussion and Future Work}
\label{discuss-sec}
\label{rel-work-sec}

We induced event pairs using several methods from previous work with
similar aims but widely different problem formulations and evaluation
methods. We used a verb-rich film scene corpus where events are
normally narrated in temporal ordered.  We used Mechanical Turk to
evaluate the learned pairs of {\sc contingent} events using human
perceptions.  In the first stage drawing on previous measures of
distributional co-occurrence, we achieved an overall average accuracy
of around 70\%, over a 50\% baseline. We then implemented a novel
method of defining narrative sequence patterns using the Google Search
API, and used web counts to further refine our estimates of the
contingency of the learned event pairs. This increased the overall
average accuracy to around 77\%, which is 27\% above the baseline.
Our results indicate that the use of web search counts increases the
average accuracy of our Causal Potential-based method to 85.64\% as
compared to an average accuracy of 75.15\% without web search.  To our
knowledge this is the highest accuracy achieved in tasks of this kind
to date.

Previous work on recognition of the PDTB {\sc contingent} relation has
used both supervised and unsupervised learning, and evaluation
typically measures precision and recall against a PDTB annotated
corpus
\cite{DoChaRo11,Pitleretal09,zhou-etal-coling10,Chiarcos12,Louisetal10}.
We use an unsupervised approach and measure accuracy using human
perceptions.  Other work by Girju and her students defined a measure
called causal potential and then used film screen plays to learn a
knowledge base of causal pairs of events. They evaluate the pairs by
asking two trained human annotators to label whether occurrences of
those pairs in their corpus are causally related
\cite{BeamerGirju09,RiazGirju10}.  We also make use of their causal
potential measure.  Work on commonsense causal reasoning aims to learn
causal relations beween pairs of events using a range of methods
applied to a large corpus of weblog narratives
\cite{Gordonetal11,GordonSwanson09,Manshadietal08}.  One form of
evaluation aimed to predict the last event in a sequence
\cite{Manshadietal08}, while more recent work uses the learned pairs
to improve performance on the COPA SEMEVAL task \cite{Gordonetal11}.

Related work on {\sc script learning} induces {\it likely sequences of
  temporally ordered events} in news, rather than {\sc contingency} or
{\sc causality} \cite{ChambersJurafsky08,ChambersJurafsky09}.
Chambers \& Jurafsky also evaluate against a corpus of existing
documents, by leaving one event out of a document (news story), and
then testing the system's ability to predict the missing event.  To
our knowledge, our method is the first to augment distributional
semantics measures from a corpus with web search data. We are also the
first to evaluate the learned event pairs with a human perceptual
evaluation with native speakers. 

We hypothesize that there are several advantages to our method: (1)
events in the same genre tend to be more similar than events across
genres, so less data is needed to estimate co-occurrence; (2) film
scenes are typically narrated via simple tenses in the correct
temporal order, which allows the ordering of events to contribute to
estimates of the {\sc contingency} relation; (3) The web counts focus
on validating event pairs already deemed to be likely to be {\sc
  contingent} in the smaller, more controlled, film scence corpus.

Our work capitalizes on event sequences narrated in temporal order 
as a cue to causality. We expect this approach to generalize to other 
domains where these properties hold, such as fables, personal stories 
and news articles. We do not expect this technique to generalize without 
further refinements to genres frequently told out of temporal order or 
when events are not mentioned consecutively in the text, for example 
in certain types of fiction.

In future work we want to explore in more detail the differences in
performance of the different contingency measures. For example,
previous work would suggest that the the higher the measure is, the
more likely the two events are to be contingent on one another. To
date, while we have only tested the top 100, we have not found that
the bottom set of 20 are less accurate than the top set of 20. This
could be due to corpus size, or the measures themselves, or noise from
parser accuracy etc.  As shown in Table~\ref{eval_all_methods} web
search refinement is able to eliminate most noise in event pairs,
but we would still aim to achieve a better understanding of the
circumstances which lead particular methods to work better. 

In future work we also want to explore ways of inducing larger
event structures than event pairs, such as the causal chains, scripts,
or narrative schemas of previous work.

\section*{Acknowledgments}

We would like to thank Yan Li for setting up automatic search
query. We also thank members of NLDS for their discussions and suggestions,
especially Stephanie Lukin, Rob Abbort, and Grace Lin.

\bibliographystyle{naaclhlt2013}

\end{document}